\documentclass[10pt,twocolumn,letterpaper]{article}

\usepackage[final]{cvpr}      
\usepackage{adjustbox}
\usepackage{multirow}
\usepackage[table]{xcolor}
\usepackage{float}

\definecolor{cvprblue}{rgb}{0.21,0.49,0.74}
\usepackage[pagebackref,breaklinks,colorlinks,allcolors=cvprblue]{hyperref}

\def\paperID{*****} 
\def\confName{CVPR}
\def\confYear{2026}

\title{SplatBright: Generalizable Low-Light Scene Reconstruction from Sparse Views via Physically-Guided Gaussian Enhancement}

\author{
Yue Wen$^{1}$,
Liang Song$^{2}$,
Hesheng Wang$^{1}$\textsuperscript{*} \\
\small{$^1$Shanghai Jiao Tong University, $^2$China DXR Technology CO.,Ltd, \textsuperscript{*}Corresponding Author}
}

\begin{document}

\twocolumn[{
\renewcommand\twocolumn[1][]{#1}

\maketitle

\begin{center}
    \centering
    \vspace{-0.5cm}
    \includegraphics[width=\linewidth]{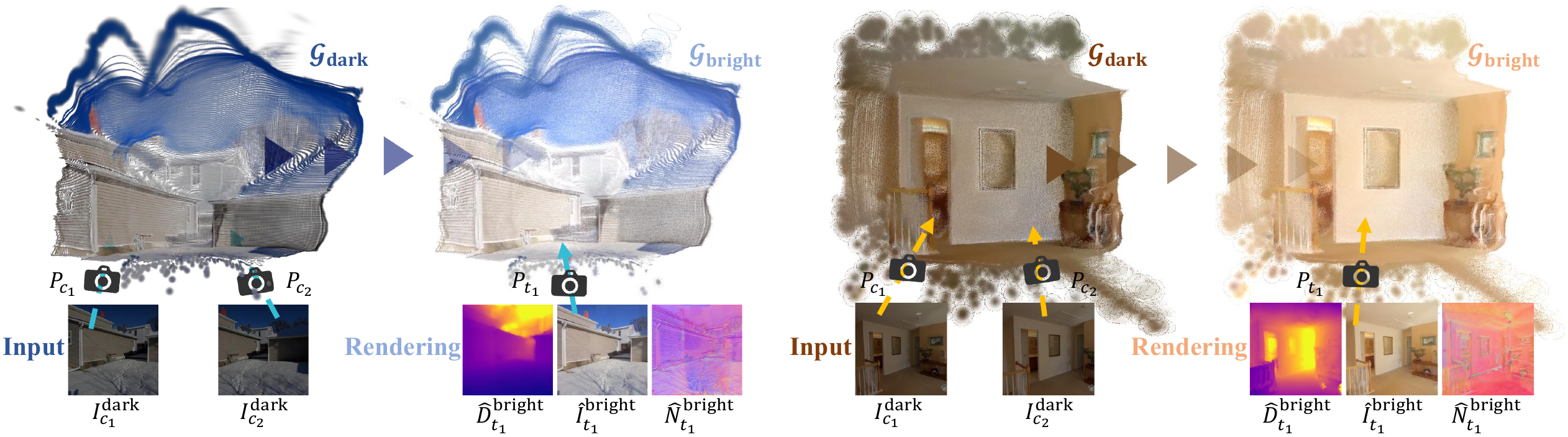}
    \vspace{-0.5cm}
    \captionof{figure}{From sparse low-light views, our method reconstructs a dark 3D Gaussian field and enhances it into a consistent normal-light field, achieving consistent illumination and geometry in novel view synthesis with depth and normal maps without per-scene training.
    } 
    \label{fig:first}
\end{center}
}]

\begin{abstract}

Low-light 3D reconstruction from sparse views remains challenging due to exposure imbalance and degraded color fidelity. While existing methods struggle with view inconsistency and require per-scene training, we propose SplatBright, which is, to our knowledge, the first generalizable 3D Gaussian framework for joint low-light enhancement and reconstruction from sparse sRGB inputs. Our key idea is to integrate physically guided illumination modeling with geometry–appearance decoupling for consistent low-light reconstruction. Specifically, we adopt a dual-branch predictor that provides stable geometric initialization of 3D Gaussian parameters. On the appearance side, illumination consistency leverages frequency priors to enable controllable and cross-view coherent lighting, while an appearance refinement module further separates illumination, material, and view-dependent cues to recover fine texture. To tackle the lack of large-scale geometrically consistent paired data, we synthesize dark views via a physics-based camera model for training.
Extensive experiments on public and self-collected datasets demonstrate that SplatBright achieves superior novel view synthesis, cross-view consistency, and better generalization to unseen low-light scenes compared with both 2D and 3D methods.

\end{abstract}

\section{Introduction}
\label{sec:intro}

Low-light scenes remain a key challenge for 3D reconstruction.
Reduced illumination lowers contrast, distorts color, and amplifies noise, hindering recovery of geometry and appearance.
Such degradation affects localization, mapping, and rendering in robotics, autonomous driving, and immersive applications, calling for a framework that directly restores accurate structure and illumination under low light.

Conventional low-light enhancement methods improve visibility in 2D images.
Classical approaches such as histogram equalization and Retinex decomposition~\cite{pizer1987adaptive,land1971lightness,guo2016lime} enhance brightness with handcrafted priors but often amplify noise and color shifts.
Deep models~\cite{chen2018deep,zhang2019kindling,Chen2018Retinex,jiang2021enlightengan} restore exposure and color using paired or adversarial training, producing visually pleasing but view-inconsistent results, since they operate per frame without enforcing geometric or photometric consistency across views.

Neural scene representations unify geometry and appearance modeling.
NeRF~\cite{mildenhall2021nerf} and its variants~\cite{mildenhall2022nerf,verbin2024ref,huang2022hdr,wang2023lighting,martin2021nerf,cui2024aleth} achieve photorealistic rendering under complex lighting and address low-light scenes.
However, these methods require dense inputs and expensive optimization.
Explicit representations such as 3D Gaussian Splatting (3DGS)~\cite{kerbl20233d} enable efficient differentiable rendering via Gaussian primitives.
Recent extensions~\cite{kulhanek2024wildgaussians,jiang2024gaussianshader,ye2024gaussian} improve fidelity and robustness but still rely on dense views, scene-specific tuning, or RAW data.
This motivates a 3D Gaussian framework unifying illumination and sparse-view reconstruction for consistent low-light recovery from sRGB inputs without per-scene training.

Building such a framework remains challenging.
Firstly, real datasets rarely offer geometry-aligned multi-view images with varying exposures, hindering the learning of illumination attenuation and cross-view consistency.
Secondly, sparse views weaken geometric cues and make appearance estimation illumination-sensitive, while uniform color adjustments in 3D space amplify view-dependent differences.
Finally, noise and unstable brightness destabilize optimization and degrade reconstruction quality.

To overcome these challenges, we propose \textbf{SplatBright}, the first 3D Gaussian framework that jointly enhances illumination and reconstructs geometry from sparse-view images, generalizing to unseen low-light scenes without per-scene optimization.
Firstly, a camera-inspired darkening model synthesizes controllable dark–normal pairs for training, complemented by a self-collected real world multi-exposure dataset for evaluation.
Secondly, a dual-branch Gaussian predictor then decouples geometry and appearance for stable modeling, while two frequency-guided modules enable hierarchical appearance enhancement:
the Illumination Consistency Module (ICM) performs controllable global illumination adjustment and exposure alignment via frequency-guided cross attention (FGCA) and style modulation;
and the Appearance Refinement Module (ARM) employs windowed 3D cross-attention across illumination, material, and view cues to restore local texture and reflection details.
Finally, a progressive training strategy sequentially optimizes geometry, global illumination, and local appearance, achieving physically consistent, detail-preserving reconstruction under low-light and sparse-view conditions. 
At inference time, SplatBright enables controllable relighting by adjusting the illumination style in unseen scenes. Our main contributions are summarized as follows:

• We present SplatBright, which is, to our knowledge, the first generalizable 3D Gaussian framework for low-light reconstruction from sparse views, trained on physically inspired dark–normal pairs.
Through geometry–appearance decoupling and progressive optimization, it enhances illumination and structure from sRGB inputs.

• We design a two-stage appearance model.
The illumination consistency module offers controllable lighting and cross-view consistency via wavelet-based style modulation, while the appearance refinement module refines each Gaussian appearance using illumination, material, and view cues.

• Extensive experiments on synthetic and real datasets, including public and self-collected datasets, show that SplatBright outperforms previous 2D and 3D methods with superior reconstruction quality and generalization.
\section{Related Work}
\subsection{2D Low-light Image Enhancement}
Low-light image enhancement has evolved from traditional algorithms to data-driven deep models.
Classical approaches based on histogram equalization, Retinex decomposition, and illumination correction~\cite{pizer1987adaptive,ibrahim2007brightness,land1971lightness,jobson1997multiscale,guo2016lime,fu2016weighted} are efficient but often amplify noise and distort color.
Deep networks~\cite{Chen2018Retinex,zhang2019kindling,chen2018deep,liu2021retinex,chen2021hinet,zamir2022restormer,liang2021swinir} learn paired mappings for exposure and color recovery, achieving high fidelity yet requiring aligned data.
To relax data dependence, unsupervised methods~\cite{Zero-DCE,jiang2021enlightengan,wang2022low,liu2020real,ma2022toward,zhu2017unpaired,jiang2023low} leverage self-regularization, adversarial learning, or diffusion priors for label-free enhancement.
Despite progress, most 2D models still enhance frames independently, lacking the geometric and photometric consistency for multi-view reconstruction.

\subsection{3D-aware Low-light Reconstruction}
Neural scene representations enable unified modeling for multi-view rendering.
NeRF~\cite{mildenhall2021nerf} and its variants~\cite{verbin2024ref,huang2022hdr,wang2023lighting,qu2024lush,mildenhall2022nerf} achieve photorealistic rendering under diverse lighting and materials.
NeRF-W~\cite{martin2021nerf} handles in-the-wild illumination with per-image embeddings, while Aleth-NeRF~\cite{cui2024aleth} models light attenuation via a concealing field.
However, volumetric NeRFs require costly optimization.
To address this, 3DGS~\cite{kerbl20233d} use explicit Gaussian primitives for faster optimization and real-time rendering.
Recent works~\cite{kulhanek2024wildgaussians,jiang2024gaussianshader,sun2025ll,hdr_gs,singh24_hdrsplat,wang2024cinematic,zhang2024gaussian} further enhance fidelity and robustness under complex lighting.
DarkGS~\cite{zhang2024darkgs} integrates neural illumination fields with camera–light calibration, Luminance-GS~\cite{cui2025luminance} aligns per-view exposures via tone mapping, and Gaussian-DK~\cite{ye2024gaussian} embeds camera parameters to decouple radiance from sensor response.
Despite these progress, most methods still require dense views, per-scene optimization, or RAW inputs, hindering generalization to sparse-view reconstruction from sRGB data.

\begin{figure*}[t]
  \centering
  \includegraphics[width=0.98\linewidth]{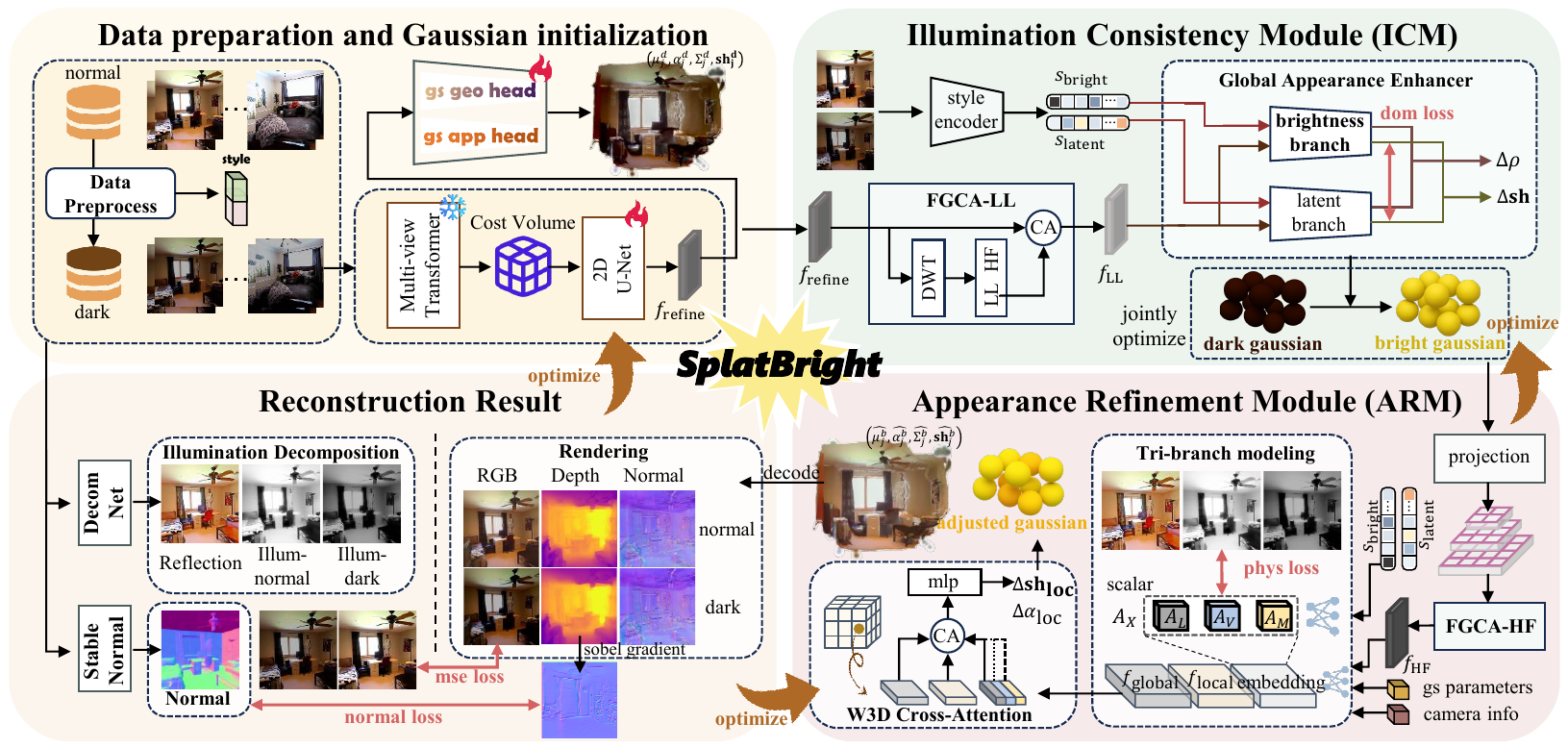}
  \vspace{-3mm}
  \caption{
  Overview of \textbf{SplatBright}.
  The pipeline includes data preprocessing, multi-view feature extraction, and geometry–appearance dual-head Gaussian initialization with normal-guided geometry optimization. With the FGCA module, ICM performs controllable lighting adjustment, while ARM models illumination, material, and view branches to enhance details for physically consistent relighting.
  }
  \label{fig:method}
  \vspace{-5mm}
\end{figure*}

\subsection{Feed-forward NeRF and 3DGS Models}
Feed-forward paradigms remove per-scene optimization by directly predicting 3D representations from sparse views.
For implicit fields, pixelNeRF~\cite{yu2021pixelnerf} generalizes to few views with pixel-aligned features, while \cite{chen2021mvsnerf,wang2021ibrnet,du2023cross,sajjadi2022scene} improve consistency via cost volumes but are costly.
Explicit 3D Gaussians enable efficient feed-forward reconstruction.
PixelSplat~\cite{charatan2024pixelsplat} employs epipolar transformers for probabilistic Gaussian sampling, and~\cite{szymanowicz2024splatter,szymanowicz2025flash3d,chen2024lara,wewer2024latentsplat,zheng2024gps} extend to single-view or stereo settings.
MVSplat~\cite{chen2024mvsplat} builds a differentiable multi-view cost volume to infer Gaussian attributes.
While recent methods~\cite{chang2025meshsplat,wan2025s2gaussian} advance sparse-view reconstruction, they ignore low-light degradation.
Our method jointly models scene geometry and illumination within a physically grounded Gaussian framework to achieve low-light reconstruction from sparse views.

\section{Method}
\label{sec:method}


As illustrated in \cref{fig:first}, our goal is to reconstruct a physically consistent 3D Gaussian field from sparse low-light inputs and generate enhanced renderings in novel views. 
Given two low-light input images $\mathcal{I}=\{I^{\text{dark}}_{c_1}, I^{\text{dark}}_{c_2}\}$ with corresponding camera matrices $\mathcal{P}=\{P_{c_1},P_{c_2}\}$, the framework predicts a unified set of 3D Gaussian primitives $\mathcal{G}=\{g_j=(\mu_j,\Sigma_j,\alpha_j,\mathbf{sh}_j)\}_{j=1}^{M}$, 
where $\mu_j$, $\Sigma_j$, $\alpha_j$, and $sh_j$ denote the position, covariance, opacity, and spherical harmonics, respectively. 
Here, $M=H\times W\times K$ represents the total number of pixel-aligned Gaussians determined by the input image resolution and the number of source views $K$. 
The reconstructed low-light field $\mathcal{G}_{\text{dark}}$ is subsequently enhanced into a normal-light field $\mathcal{G}_{\text{bright}}$, 
enabling rendering of photometrically consistent novel views
$
\hat{\mathcal{I}}^{\text{bright}} = \{\hat{I}_{t_i}\}_{i=1}^{N},
$
under known target camera poses
$
\mathcal{P}_t = \{P_{t_i}\}_{i=1}^{N},
$
where $N$ denotes the number of target views.
The pipeline is illustrated in \cref{fig:method}.

\subsection{Data preparation and Gaussian initialization}

\begin{figure}[t]
  \centering
  \includegraphics[width=\linewidth]{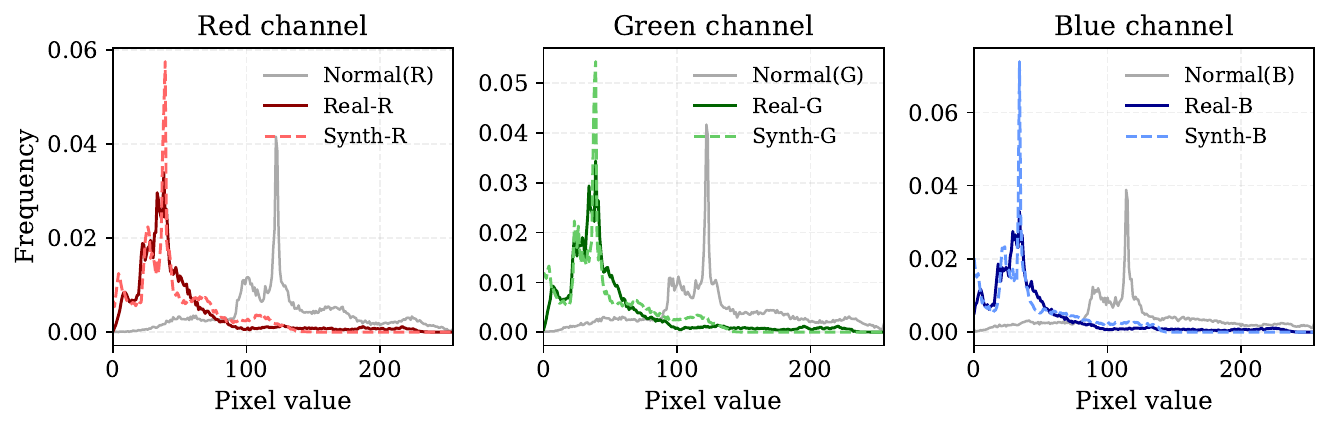}
  \vspace{-6mm}
  \caption{
    Comparison of per RGB channel intensity statistics.
  }
  \label{fig:rgb_stats}
  \vspace{-5mm}
\end{figure}

\noindent\textbf{Dark data generation.}
\label{sec:darkdata}
Collecting multi-view low-light datasets with consistent geometry is highly challenging in practice.  
To overcome this, we synthesize dark-normal pairs $(I_\text{dark}, I_\text{normal})$ on the RealEstate10K (RE10K)~\cite{10.1145/3197517.3201323} dataset, which provides geometry-aligned normal light images, by applying a physically inspired degradation process that mimics how cameras perceive dim scenes:
\begin{equation}
I_\text{dark} = \mathcal{T}_{\text{dark}}(I_\text{normal}; d),
\end{equation}
where $\mathcal{T}_{\text{dark}}$ applies exposure drop, ISP-tone compression, and chroma suppression, controlled by $d$.  
However, darkening tends to wash out bright sky regions into dull gray.  
To prevent unrealistic artifacts, we use a soft sky mask $M_\text{sky}$:
\begin{equation}
I_\text{dark} \leftarrow (1-M_\text{sky}) I_\text{dark} + M_\text{sky} I_\text{normal}.
\end{equation}
Our synthetic dark images match real low-light RGB statistics in~\cref{fig:rgb_stats}, confirming the realism of the degradation model. By varying darkness levels $(d_\text{low}, d_\text{high})$, we create controllable supervision pairs, allowing the model to learn robust illumination adaptation.

\noindent\textbf{Multi-view depth estimation.}
We first recover geometry via multi-view disparity estimation, convert disparity to depth, and unproject into 3D space to initialize Gaussian centers $\mu_j^d$ and coarse opacity $\alpha_j^d$.  
A transformer–U-Net backbone extracts geometric features. To stabilize depth reasoning under low-light conditions, the low-level encoder filters are kept fixed while the refinement layers remain learnable, producing latent features $F=\phi_{\text{gs}}(f_{\text{refine}})$, where $f_{\text{refine}}$ denote the U-Net feature map from the depth branch.
Considering the distribution discrepancy between geometry and appearance, 
we design a decoupled dual-head Gaussian predictor for stable low-light modeling:
\begin{equation}
\mathbf{g}^{\text{geo}} = \psi_{\text{geo}}(F), \quad
\mathbf{g}^{\text{app}} = \psi_{\text{app}}(F),
\end{equation}
which are concatenated as $\mathbf{g}^{d} = \operatorname{concat}(\mathbf{g}^{\text{geo}},\, \mathbf{g}^{\text{app}}) \in \mathbb{R}^{C \times H \times W}$,
where $C$ denotes the total number of geometric and appearance parameters.  
The geometry branch refines gaussian centers, scales, and opacities to obtain anisotropic covariances $\Sigma_j^d$,  
while the appearance branch predicts spherical harmonics(SH) $\mathbf{sh}_j^d$ to encode intrinsic color and view-dependent illumination.  
The resulting tensor $\mathbf{g}^d$ is finally decoded into a set of gaussian primitives:
\begin{equation}
\mathcal{G}_{\text{dark}} = \{ (\mu_j^d, \alpha_j^d, \Sigma_j^d, \mathbf{sh}_j^d) \}_{j=1}^{H \times W \times K}.
\end{equation}

\begin{figure}[t]
  \centering
   \includegraphics[width=1.0\linewidth]{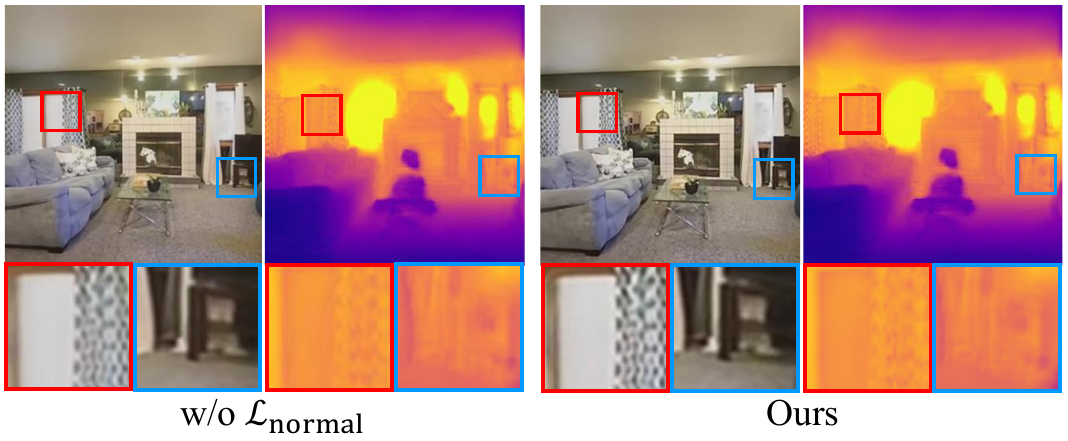}
   \vspace{-7mm}
   \caption{Normal supervision $\mathcal{L}_{\text{normal}}$ in the geometry stage improves texture sharpness and depth estimation.}
   \label{fig:ablation2}
   \vspace{-3mm}
\end{figure}

\subsection{Illumination Consistency Module (ICM)}
\label{sec:icm}

\begin{figure}[t]
  \centering
   \includegraphics[width=1.0\linewidth]{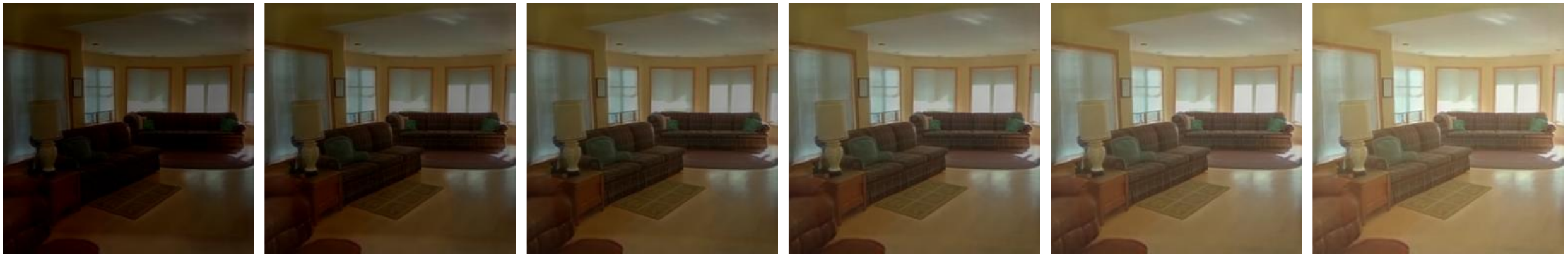}
   \vspace{-7mm}
   \caption{Inference results using different brightness values $s_{\text{bright}}$ ranging from $-1.0$ to $1.5$..}
   \label{fig:light}
   \vspace{-6mm}
\end{figure}

\begin{figure*}[t]
  \centering
  \begin{subfigure}{0.48\linewidth}
    \centering
    \includegraphics[width=\linewidth]{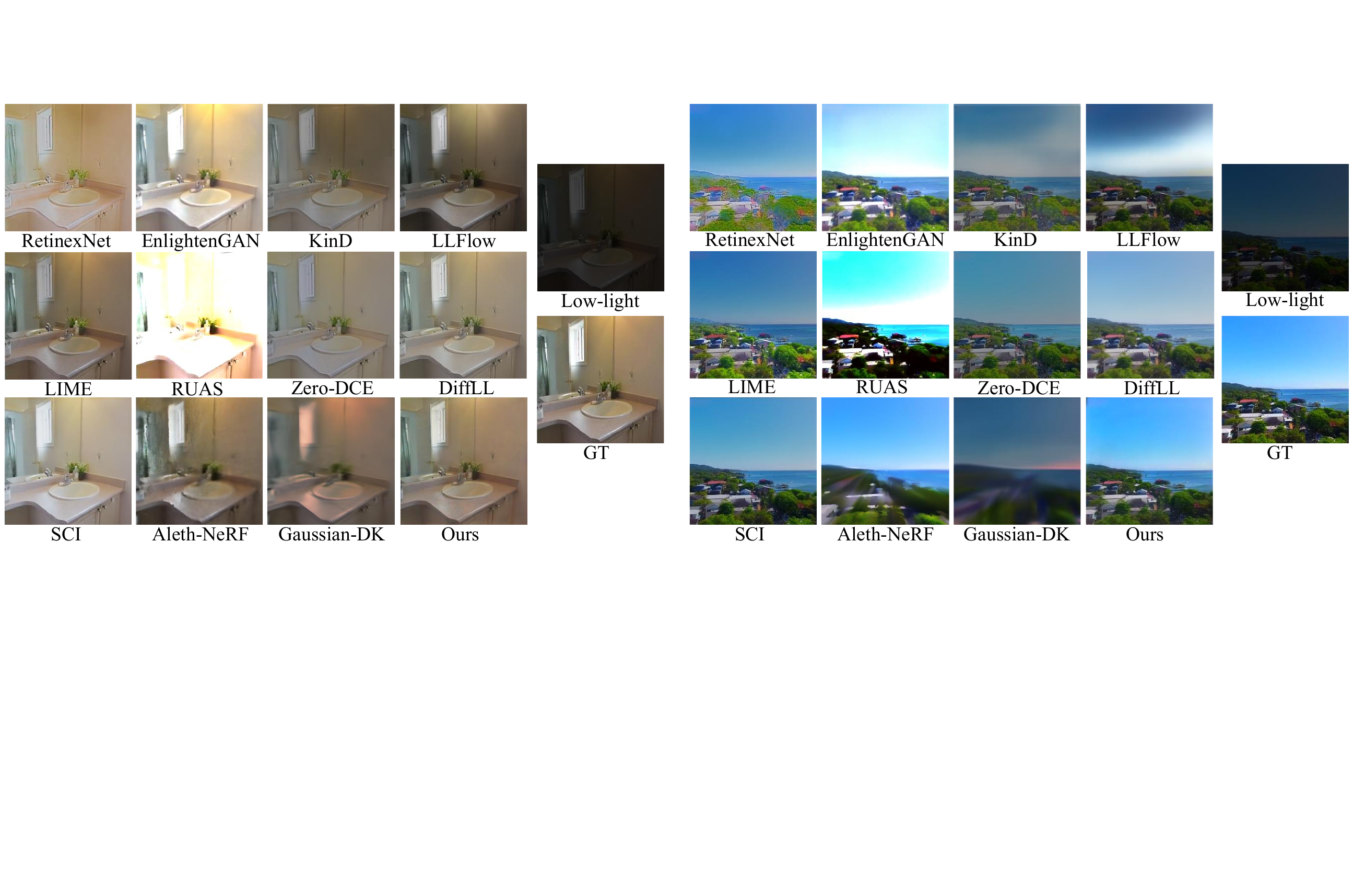}
    \caption{RE10K}
    \label{fig:re10k}
    \vspace{-4mm}
  \end{subfigure}
  \hfill
  \begin{subfigure}{0.48\linewidth}
    \centering
    \includegraphics[width=\linewidth]{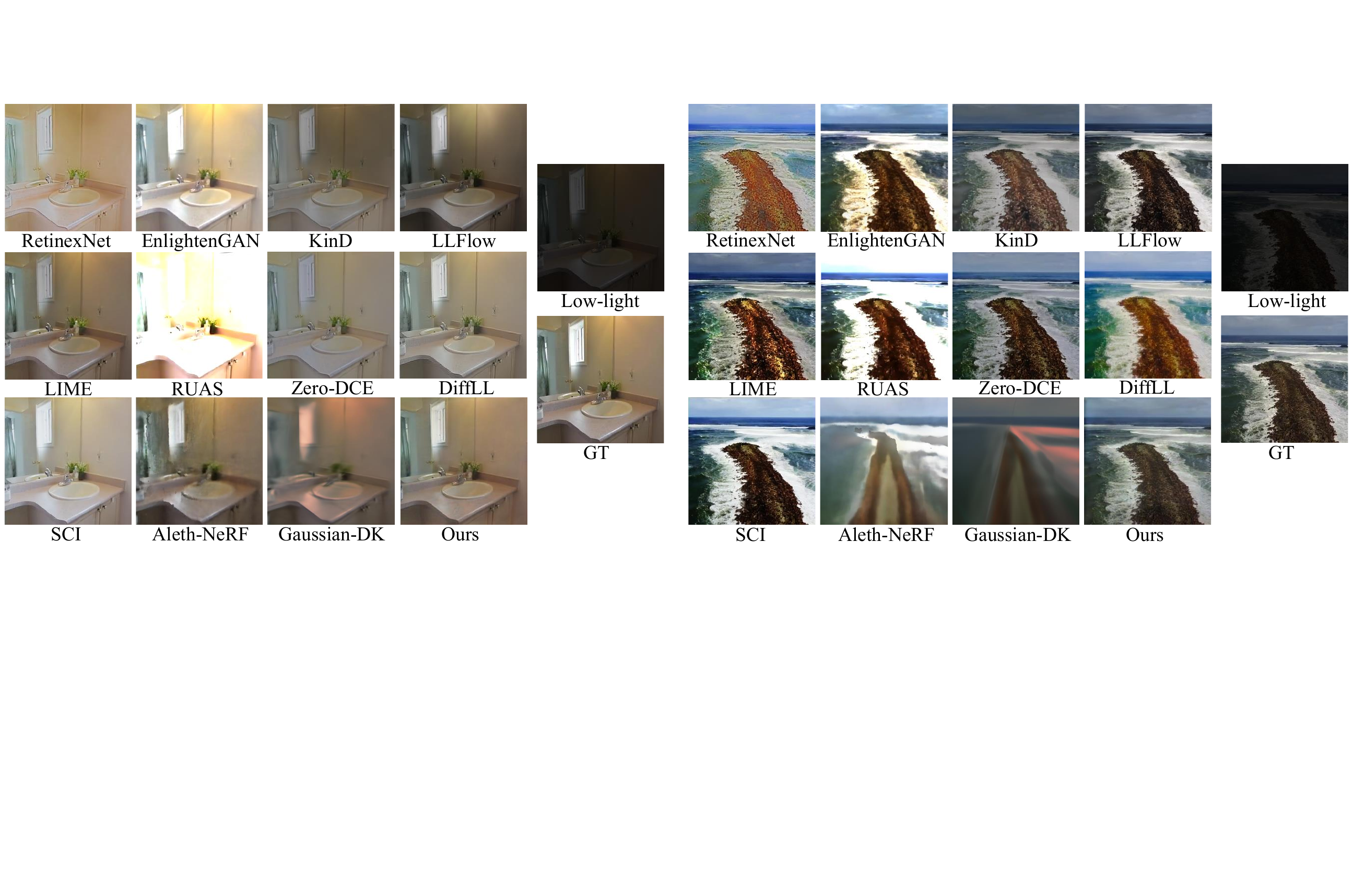}
    \caption{RE10K $\rightarrow$ ACID}
    \label{fig:acid}
    \vspace{-4mm}
  \end{subfigure}

  \caption{
    Visualization of novel view synthesis results.
    (a) Results on the RE10K dataset, tested after training.  
    (b) Generalization results on ACID via direct cross-dataset inference (RE10K $\rightarrow$ ACID).
  }
  \label{fig:re10k_acid}
  \vspace{-6mm}
\end{figure*}

After initializing the dark Gaussian field, direct brightness scaling often amplifies noise and causes grayish bias.
To ensure global consistency, we design an Illumination Consistency Module (ICM) that integrates low-frequency information with style-modulated illumination–color disentanglement,
enabling global gaussian parameters adjustment and scene-level controllable brightness prediction.

\noindent\textbf{Frequency-guided cross attention (FGCA).}
To obtain a cross-view consistent global signal, we apply a 2D discrete wavelet transform (DWT) to the refined 2D feature \(f_{\text{refine}} \in \mathbb{R}^{C \times H \times W}\),
decomposing it into four frequency subbands \([LL,\,HL,\,LH,\,HH] = f_{\mathrm{DWT}}(f_{\text{refine}})\),
where the component \(LL \in \mathbb{R}^{C \times \tfrac{H}{2} \times \tfrac{W}{2}}\) captures low-frequency global smooth information and is less sensitive to noise.  
We downsample \(f_{\text{refine}}\) to obtain \(\tilde{f}_{\mathrm{refine}}\) and perform cross-attention to extract a robust global descriptor:
\begin{equation}
f_{\mathrm{low}} =
\mathrm{softmax}\!\left(
\frac{\phi_q(\tilde{f}_{\text{refine}})\phi_k(LL)^{\top}}{\sqrt{C'}}
\right)\phi_v(LL),
\label{eq:dwt}
\end{equation}
where \(\phi_q, \phi_k, \phi_v\) denote the learnable projections for the query, key, and value, 
and \(C'\) represents the channel dimension used to normalize the attention logits.
The attended feature is upsampled and added back to the original feature in a residual manner, 
forming feature \(f_{\mathrm{LL}}\) constrained by low-frequency priors.

\noindent\textbf{Global appearance adjustment.}
A lightweight dark style predictor encodes the low-light and normal inputs $I$:
\begin{equation}
[s_{\text{bright}}, s_{\text{latent}}] = f_{\text{style}}(I),
\end{equation}
where $s_{\text{bright}}$ represents an explicit brightness difference factor and $s_{\text{latent}}$ captures other degradations such as contrast and color shifts. 
These interpretable style cues enable explicit illumination enhancement across views.

Guided by the frequency-aware feature \(f_{\mathrm{LL}}\) and the style code \(s=[s_{\text{bright}}, s_{\text{latent}}]\), 
a dual-branch global enhancer modulates the SH coefficients and opacity:
\begin{equation}
\begin{aligned}
\Delta \mathbf{sh} &= f_{\text{b}}(s_{\text{bright}}, f_{\text{refine}}) + f_{\text{l}}(s_{\text{latent}}, f_{\text{refine}}), \\
\Delta \rho &= g_{\text{b}}(s_{\text{bright}}, f_{\text{refine}}) + g_{\text{l}}(s_{\text{latent}}, f_{\text{refine}}),
\end{aligned}
\end{equation}
where \(f_{\text{b}}, f_{\text{l}}, g_{\text{b}}, g_{\text{l}}\) denote the brightness and latent modulation branches. 
The brightness term controls global exposure (~\cref{fig:light}), while the latent term refines tone and contrast.  

SH coefficients are updated in a residual form, and opacity follows an exponential decay of radiance with optical density, analogous to light attenuation in hazy scenes:
\begin{equation}
\tilde{\rho} = \rho \exp(-\gamma_\rho \, \Delta \rho), \qquad
\tilde{\mathbf{sh}} = \mathbf{sh} + \Delta \mathbf{sh},
\end{equation}
where $\rho$ is the per-gaussian density before opacity mapping, and the decay factor $\gamma_\rho$ controls the sensitivity of density adjustment. A dominance regularization
\begin{equation}
\mathcal{L}_{\text{dom}} = 
\lambda_{\text{dom}} 
\frac{\|\Delta \mathbf{sh}^{(l)}\|_1}{\|\Delta \mathbf{sh}^{(b)}\|_1+\epsilon}
\end{equation}
ensures interpretable control, letting \(s_{\text{bright}}\) dominate exposure, where \(\lambda_{\text{dom}}\) sets the weight and \(\epsilon\) is a small constant.






\subsection{Appearance Refinement Module (ARM)}
\label{sec:arm}
While the global enhancer corrects scene-level illumination, real scenes still exhibit local variations such as uneven lighting, material-dependent reflectance, and view-dependent highlights. 
We therefore introduce an Appearance Refinement Module (ARM) to perform per-gaussian corrections on top of the global result.

\noindent\textbf{High-level feature construction.}
High-frequency parts \((HL, LH, HH)\) are processed by FGCA similar to Eq.~\eqref{eq:dwt} to obtain \(f_{\mathrm{HF}}\).
A multi-scale pyramid \(\{F_i^{(l)}(f_{\mathrm{HF}})\}\) is then built for each view, 
where \(F_i^{(l)}(\cdot)\) denotes the feature map of view \(i\) at level \(l\).
Each gaussian center \(x_j\) is projected to all views, and multi-view features are bilinearly sampled as
\begin{equation}
f_{\text{sample}} = \Phi_{\text{sample}}(x_j, \{F_i^{(l)}(f_{\mathrm{HF}})\}_{i,l}),
\end{equation}
We build two descriptors for each gaussian: 
the local descriptor \(f_{\text{local}}\) merges $f_{\text{sample}}$ with multi-view and geometric cues, 
while the global descriptor \(f_{\text{global}}\) includes SH, opacity, and style to represent illumination. 
These features capture texture and shading for robust local texture refinement.

\begin{table}[t]
\caption{
Novel view synthesis results of RE10K dataset.
Best results are red, second-best are blue.
* denotes MVSplat.
} 
\vspace{-3mm}
\label{tab:re10k}
\centering
\renewcommand\arraystretch{1.8}
\begin{adjustbox}{max width = 1\linewidth}
\begin{tabular}{l|c|l|c}
\toprule
\toprule
Method & PSNR / SSIM / LPIPS & Method & PSNR / SSIM / LPIPS \\ 
\midrule
\midrule
SCI + *            & 18.23 / \textbf{\textcolor{blue}{0.809}} / 0.189 & * + SCI          & 18.13 / 0.804 / 0.228 \\ \hline
Zero-DCE + *       & 15.50 / 0.761 / 0.259 & * + Zero-DCE     & 12.70 / 0.665 / 0.357 \\ \hline
DLL + *            & 17.52 / 0.780 / 0.214 & * + DLL          & 17.29 / 0.742 / 0.293 \\ \hline
Retinex + *        & 15.11 / 0.718 / 0.243 & * + Retinex      & 14.97 / 0.714 / 0.254 \\ \hline
LIME + *           & 15.10 / 0.781 / \textbf{\textcolor{blue}{0.189}} & * + LIME         & 14.88 / 0.767 / 0.218 \\ \hline
KinD + *           & 14.20 / 0.716 / 0.290 & * + KinD         & 13.66 / 0.708 / 0.306 \\ \hline
LLFlow + *         & 16.48 / 0.787 / 0.208 & * + LLFlow       & 16.31 / 0.777 / 0.218 \\ \hline
EnlightenGAN + *   & 15.57 / 0.783 / 0.209 & * + EnlightenGAN & 15.30 / 0.770 / 0.228 \\ \hline
RUAS + *           & 11.49 / 0.632 / 0.473 & * + RUAS         & 11.50 / 0.624 / 0.477 \\ \hline
Gaussian-DK      & 14.54 / 0.632 / 0.484 & Aleth-NeRF         & \textbf{\textcolor{blue}{20.74}} / 0.721 / 0.404 \\ \hline
\textbf{Ours}      & \textbf{\textcolor{red}{21.43}} / \textbf{\textcolor{red}{0.815}} /  \textbf{\textcolor{red}{0.168}} 
                   & MVSplat &  6.72 / 0.264 / 0.348 \\ 
\bottomrule
\bottomrule
\end{tabular}
\end{adjustbox}
\vspace{-5mm}
\end{table}

\begin{table}[t]
\caption{
Generalization on the ACID dataset (RE10K $\rightarrow$ ACID). Retrain Gaussian-DK and Aleth-NeRF. * denotes MVSplat. 
} 
\vspace{-3mm}
\label{tab:acid}
\centering
\renewcommand\arraystretch{1.8}
\begin{adjustbox}{max width = 1\linewidth}
\begin{tabular}{l|c|l|c}
\toprule
\toprule
Method & PSNR / SSIM / LPIPS & Method & PSNR / SSIM / LPIPS \\ 
\midrule
\midrule
SCI + *            & 17.52 / 0.755 / 0.232 & * + SCI          & 17.52 / 0.755 / 0.232 \\ \hline
Zero-DCE + *       & 19.27 / 0.792 / \textbf{\textcolor{blue}{0.202}} & * + Zero-DCE     & 19.38 / \textbf{\textcolor{blue}{0.794}} / 0.216 \\ \hline
DLL + *            & 19.84 / 0.779 / 0.244 & * + DLL          & \textbf{\textcolor{blue}{19.94}} / 0.753 / 0.278 \\ \hline
Retinex + *        & 16.77 / 0.732 / 0.262 & * + Retinex      & 16.77 / 0.732 / 0.262 \\ \hline
LIME + *           & 17.97 / 0.708 / 0.223 & * + LIME         & 17.31 / 0.709 / 0.236 \\ \hline
KinD + *           & 16.72 / 0.777 / 0.276 & * + KinD         & 16.72 / 0.777 / 0.276 \\ \hline
LLFlow + *         & 17.95 / 0.772 / 0.221 & * + LLFlow       & 17.61 / 0.770 / 0.237 \\ \hline
EnlightenGAN + *   & 16.97 / 0.788 / 0.265 & * + EnlightenGAN & 16.58 / 0.778 / 0.279 \\ \hline
RUAS + *           & 11.70 / 0.644 / 0.416 & * + RUAS         & 11.63 / 0.640 / 0.424 \\ \hline
Gaussian-DK & 11.55 / 0.482 / 0.564 & Aleth-NeRF & 15.89 / 0.521 / 0.452 \\ \hline
\textbf{Ours}    & \textbf{\textcolor{red}{22.69}} / \textbf{\textcolor{red}{0.814}} / \textbf{\textcolor{red}{0.175}} & MVSplat &  7.45 / 0.250 / 0.397 \\ 
\bottomrule
\bottomrule
\end{tabular}
\end{adjustbox}
\vspace{-5mm}
\end{table}

\noindent\textbf{Tri-branch appearance modeling.}
To capture spatially varying appearance effects, ARM models each gaussian’s appearance as three components: illumination, material, and view-dependent factors. 
They respectively capture exposure variation, surface reflectance and texture, and directional effects.  
Each branch predicts a scalar and an embedding from $f_\kappa(x)$ generated by high-level feature:
\begin{equation}
A_\kappa(x),\, e_\kappa(x)=\Phi_\kappa(f_\kappa(x)), \quad \kappa\!\in\!\{L,M,V\},
\end{equation}
where \(A_\kappa(x)\) controls modulation strength and \(e_\kappa(x)\) encodes local context. 
Illumination and material scalars are constrained by a sigmoid, while the view scalar adopts \(\tanh\) for bidirectional adjustment. 
An MLP fuses the three as 
\(A_X(x)=\Psi([A_L, A_M, A_V])\),
which, together with \(\{e_\kappa(x)\}\), drives attention-based refinement. 
The branches are supervised by normal-light illumination, reflectance, and illumination-difference maps generated by~\cite{zhang2019kindling}.

\begin{figure*}[t]
  \centering
  \includegraphics[width=0.98\linewidth]{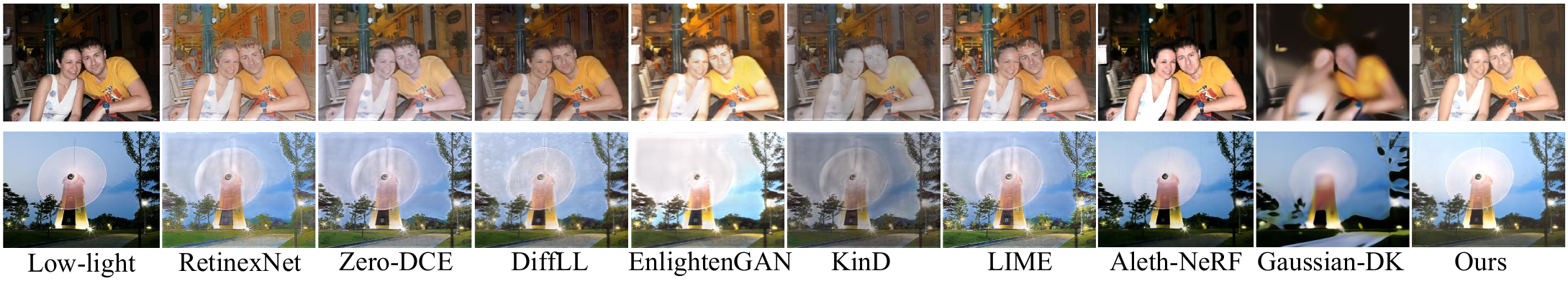}
  \vspace{-3mm}
  \caption{
  Generalization results on VV and DICM dataset via cross-dataset inference (RE10K $\rightarrow$ VV(top); RE10K $\rightarrow$ DICM(bottom)).
  }
  \label{fig:onlypic}
  \vspace{-3mm}
\end{figure*}

\begin{figure*}[t]
  \centering
  \includegraphics[width=0.98\linewidth]{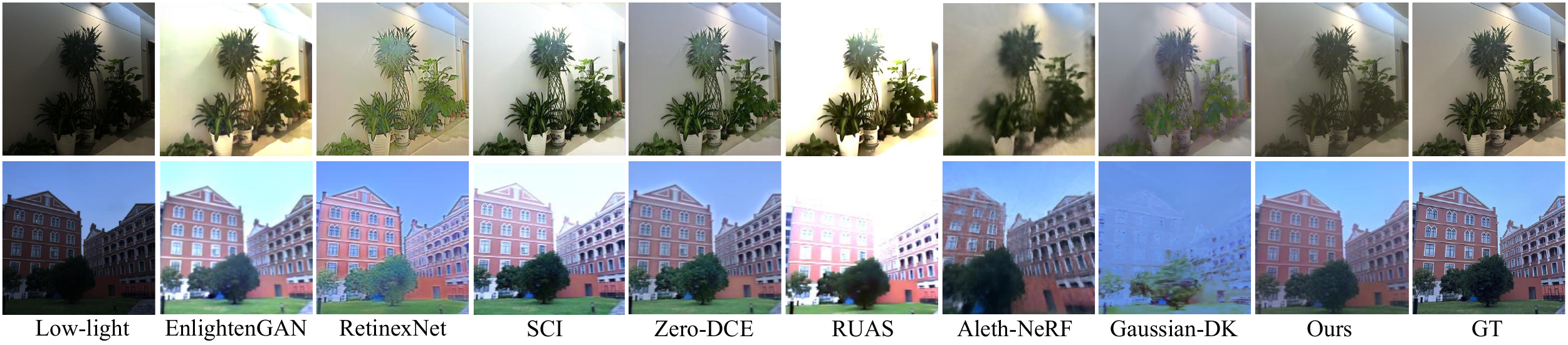}
  \vspace{-3mm}
  \caption{
  Generalization results on the L3DS dataset via cross-dataset inference (RE10K $\rightarrow$ L3DS).
  }
  \label{fig:L3DS}
  \vspace{-6mm}
\end{figure*}

We design a physically guided cross-attention where
\(f_{\text{local}}\) acts as the query,
the factor-specific embeddings $e_\kappa(x)$ serve as keys,
and \(f_{\text{global}}\) attributes provide values.
This allows each gaussian to dynamically exchange information with global attributes and retrieve only the physically relevant cues.
For efficiency, the 3D scene is divided into voxel windows,
and attention is computed within each window:
\begin{equation}
f_{\kappa} =
\mathrm{WCA}(q=f_{\text{local}},\, k=e_{\kappa},\, v=f_{\text{global}},\, \mathcal{W}_{\text{id}}),
\end{equation}
where $\kappa$ denotes illumination(low), material(mid), and view factors(high), 
and $\mathcal{W}_{\text{id}}$ specifies the window grouping. WCA represents windowed 3D cross-attention. The features are decoded into frequency-weighted residuals:
\begin{equation}
\begin{split}
 \Delta \mathbf{sh}_{\text{loc}}, \Delta \alpha_{\text{loc}} &=
[\lambda_{0} h_{\text{sh}}^{(0)}(f_{\text{low}}),\;
 \lambda_{1} h_{\text{sh}}^{(1)}(f_{\text{mid}}),\\
&\quad \lambda_{2} h_{\text{sh}}^{(2)}(f_{\text{high}})], \lambda_{\alpha}\, h_{\alpha}(f_{\text{mid}}),
\end{split}
\end{equation}
where $\lambda_i$ are weights balancing low-, mid-, and high-order corrections.
The globally attenuated density is then mapped to opacity 
$\tilde{\alpha} = \mathcal{M}(\tilde{\rho})$, and the final gaussian attributes are
\begin{equation}
\hat{\mathbf{sh}} = \tilde{\mathbf{sh}} + \Delta \mathbf{sh}_{\text{loc}}, \qquad
\hat{\alpha} = \mathrm{clip}\bigl(\tilde{\alpha} + \Delta \alpha_{\text{loc}},\, 0, 1\bigr).
\end{equation}

By combining factorized keys with windowed attention, 
this module performs local refinement where illumination, material, view, and geometry are jointly optimized.
The resulting bright Gaussian field is defined as
\begin{equation}
\mathcal{G}_{\text{bright}} = 
\{ (\hat{\mu}_j^b, \hat{\alpha}_j^b, \hat{\Sigma}_j^b, \hat{\mathbf{sh}}_j^b) \}_{j=1}^{H \times W \times K}.
\end{equation}
Given $\mathcal{G}_{\text{bright}}$, the 3DGS decoder~\cite{kerbl20233d} renders novel-view outputs, including RGB $\hat{I}$, depth $\hat{D}$, and normals $\hat{N}$.

\begin{table}[t]
\caption{
Generalization on five real-world datasets. Retrain Gaussian-DK and Aleth-NeRF. * denotes MVSplat.
}
\vspace{-2mm}
\label{tab:onlypic}
\centering
\renewcommand\arraystretch{1.4}
\setlength{\tabcolsep}{4pt}
\fontsize{12pt}{13pt}\selectfont

{\large
\begin{adjustbox}{max width=1\linewidth}
\begin{tabular}{l|c|c|c|c|c}
\toprule
\toprule
\multirow{2}{*}{Method}
& \multicolumn{5}{c}{NIQE$\downarrow$ / MUSIQ$\uparrow$} \\
& VV & NPE & DICM & LIME & MEF \\
\midrule

SCI + *           
& 4.25 / 40.52  
& 7.27 / 37.34  
& 4.70 / 42.07  
& 4.38 / 49.56  
& \underline{3.80} / 48.26 \\

Zero-DCE + *      
& 4.13 / 49.15  
& 6.32 / 45.85  
& 3.46 / 50.33  
& 4.50 / \textbf{\textcolor{blue}{53.15}}
& 3.96 / 50.82 \\

DLL + *           
& 3.89 / 43.45  
& 5.48 / 41.01  
& \textbf{\textcolor{red}{3.35}} / 45.40  
& \textbf{\textcolor{blue}{4.17}} / 44.89  
& \textbf{\textcolor{blue}{3.76}} / 45.18 \\

Retinex + *       
& 4.33 / \textbf{\textcolor{red}{56.10}}
& 5.63 / \textbf{\textcolor{red}{50.85}}
& 3.54 / \textbf{\textcolor{red}{57.70}}
& 4.87 / 52.47
& 4.19 / \textbf{\textcolor{blue}{54.80}} \\

LIME + *          
& 3.93 / 48.88
& 4.97 / 46.59
& \textbf{\textcolor{blue}{3.40}} / 52.60
& 4.55 / 51.93
& 3.84 / 50.84 \\

KinD + *          
& 4.52 / 45.16
& 6.96 / 45.65
& 3.62 / \textbf{\textcolor{blue}{52.76}}
& 4.74 / 50.95
& 4.03 / 49.69 \\

LLFlow + *        
& 4.31 / 45.31
& 6.91 / 43.94
& 3.41 / 51.34
& 4.61 / 50.54
& 3.98 / 48.06 \\

EnlightenGAN + *  
& 4.56 / 32.62
& 6.93 / 29.89
& 4.46 / 32.88
& 4.46 / 43.15
& 4.43 / 36.23 \\

RUAS + *          
& 5.56 / 30.86
& 9.41 / 31.72
& 4.28 / 40.24
& 4.78 / 41.21
& 4.34 / 42.42 \\

Aleth-NeRF 
& \textbf{\textcolor{red}{3.67}} / 21.10
& \textbf{\textcolor{blue}{4.30}} / \textbf{\textcolor{blue}{47.22}}
& 3.85 / 40.31
& 4.55 / 45.99
& 4.79 / 38.38 \\

Gaussian-DK
& 6.35 / 16.92
& 5.54 / 38.60
& 4.51 / 46.94
& 4.26 / 30.72
& 4.77 / 30.92 \\

\textbf{Ours}     
& \textbf{\textcolor{blue}{3.88}} / \textbf{\textcolor{blue}{49.18}}
& \textbf{\textcolor{red}{3.89}} / 45.46
& 3.91 / 46.70
& \textbf{\textcolor{red}{4.10}} / \textbf{\textcolor{red}{53.66}}
& \textbf{\textcolor{red}{3.59}} / \textbf{\textcolor{red}{56.59}} \\

\bottomrule
\bottomrule
\end{tabular}
\end{adjustbox}}
\vspace{-6mm}
\end{table}

\subsection{Training Strategy and Loss Functions}
\label{sec:training}

Depth errors distort geometry, and appearance correction can be unstable during optimization.  
We therefore adopt a three-stage training strategy, progressively enabling each loss as its supervision becomes reliable.

\begin{table*}[t]
\caption{
Quantitative comparison on the L3DS dataset
(“corridor”, “window”, “balcony”, “shelf”, “plant”, “campus”).
Metrics: PSNR$\uparrow$, SSIM$\uparrow$, LPIPS$\downarrow$. Retrain
Gaussian-DK and Aleth-NeRF.
* denotes MVSplat.
}
\vspace{-2mm}
\label{tab:L3DS}
\centering
\renewcommand\arraystretch{1.35}
\begin{adjustbox}{max width=\linewidth}
\begin{tabular}{lccc|ccc|ccc|ccc|ccc|ccc|ccc}
\toprule
\toprule
\multirow{2}{*}{Method}
& \multicolumn{3}{c|}{\textbf{“corridor”}}
& \multicolumn{3}{c|}{\textbf{“window”}}
& \multicolumn{3}{c|}{\textbf{“balcony”}}
& \multicolumn{3}{c|}{\textbf{“shelf”}}
& \multicolumn{3}{c|}{\textbf{“plant”}}
& \multicolumn{3}{c|}{\textbf{“campus”}}
& \multicolumn{3}{c}{\textbf{avg}} \\
\cmidrule(lr){2-4} \cmidrule(lr){5-7} \cmidrule(lr){8-10} 
\cmidrule(lr){11-13} \cmidrule(lr){14-16} \cmidrule(lr){17-19} \cmidrule(lr){20-22}
& PSNR & SSIM & LPIPS
& PSNR & SSIM & LPIPS
& PSNR & SSIM & LPIPS
& PSNR & SSIM & LPIPS
& PSNR & SSIM & LPIPS
& PSNR & SSIM & LPIPS
& PSNR & SSIM & LPIPS \\
\midrule

SCI + *           
& \textbf{\textcolor{blue}{21.36}} & 0.910 & 0.224
& 16.65 & 0.867 & 0.183
& 14.86 & 0.777 & 0.180
& 14.40 & 0.830 & 0.214
& 12.33 & 0.759 & 0.278
& 12.33 & 0.622 & 0.261
& 15.32 & 0.789 & 0.223 \\
\midrule

Zero-DCE + *      
& 20.15 & \textbf{\textcolor{red}{0.933}} & 0.236
& \textbf{\textcolor{blue}{21.23}} & \textbf{\textcolor{red}{0.908}} & 0.161
& 18.58 & 0.813 & \textbf{\textcolor{blue}{0.175}}
& \textbf{\textcolor{blue}{20.57}} & 0.839 & \textbf{\textcolor{blue}{0.209}}
& 20.37 & \textbf{\textcolor{red}{0.806}} & \textbf{\textcolor{red}{0.202}}
& \textbf{\textcolor{blue}{19.79}} & \textbf{\textcolor{red}{0.657}} & \textbf{\textcolor{red}{0.169}}
& \textbf{\textcolor{blue}{20.11}} & \textbf{\textcolor{red}{0.824}} &	\textbf{\textcolor{blue}{0.192}} \\
\midrule

DLL + *           
& 18.45 & 0.837 & 0.345
& 14.93 & 0.810 & 0.279
& \textbf{\textcolor{blue}{20.07}} & 0.794 & 0.212
& 18.90 & 0.783 & 0.262
& 13.18 & 0.765 & 0.273
& 13.42 & 0.600 & 0.255
& 16.49 & 0.770 & 0.271 \\
\midrule

Retinex + *       
& 15.48 & 0.862 & 0.280
& 16.30 & 0.821 & 0.223
& 17.38 & 0.756 & 0.217
& 14.97 & 0.758 & 0.282
& 14.83 & 0.769 & 0.224
& 13.79 & 0.597 & 0.237
& 15.46 & 0.760	& 0.244 \\
\midrule

KinD + *          
& 18.82 & 0.906 & \textbf{\textcolor{blue}{0.191}}
& 18.70 & 0.849 & \textbf{\textcolor{blue}{0.156}}
& 15.10 & 0.785 & 0.177
& 19.42 & \textbf{\textcolor{blue}{0.845}} & 0.247
& \textbf{\textcolor{blue}{20.56}} & 0.768 & 0.224
& 17.45 & 0.650 & 0.236
& 18.34 &	0.801	& 0.205 \\
\midrule

EnlightenGAN + *  
& 19.68 & 0.909 & 0.245
& 19.16 & 0.845 & 0.196
& 16.65 & \textbf{\textcolor{blue}{0.804}} & 0.227
& 14.86 & 0.832 & 0.254
& 12.56 & 0.767 & 0.322
& 12.58 & 0.639 & 0.271
& 15.91 & 0.800 &	0.253 \\
\midrule

RUAS + *          
& 12.41 & 0.785 & 0.359
& 9.27  & 0.583 & 0.440
& 9.47  & 0.512 & 0.503
& 8.77  & 0.595 & 0.541
& 9.03  & 0.655 & 0.442
& 8.87  & 0.527 & 0.435
& 9.64 & 0.610	& 0.453 \\
\midrule

Gaussian-DK   
& 17.81 & 0.745 & 0.193
& 16.66 & 0.538 & 0.318
& 16.30 & 0.462 & 0.332
& 16.36 & 0.555 & 0.367
& 16.69 & 0.596 & 0.324
& 11.69 & 0.375 & 0.629
& 15.92 & 0.545	& 0.361 \\
\midrule

Aleth-NeRF
& 12.24 & 0.448 & 0.529
& 11.09 & 0.360 & 0.612
& 10.59 & 0.311 & 0.588
& 17.59 & 0.601 & 0.610
& 17.24 & 0.573 & 0.581
& 16.05 & 0.467 & 0.616
& 14.13 & 0.460 & 0.589 \\
\midrule

\textbf{Ours}     
& \textbf{\textcolor{red}{22.84}} & \textbf{\textcolor{blue}{0.912}} & \textbf{\textcolor{red}{0.181}}
& \textbf{\textcolor{red}{21.31}} & \textbf{\textcolor{blue}{0.858}} & \textbf{\textcolor{red}{0.143}}
& \textbf{\textcolor{red}{20.35}} & \textbf{\textcolor{red}{0.806}} & \textbf{\textcolor{red}{0.162}}
& \textbf{\textcolor{red}{21.17}} & \textbf{\textcolor{red}{0.848}} & \textbf{\textcolor{red}{0.202}}
& \textbf{\textcolor{red}{20.91}} & \textbf{\textcolor{blue}{0.778}} & \textbf{\textcolor{blue}{0.215}}
& \textbf{\textcolor{red}{20.07}} & \textbf{\textcolor{blue}{0.636}} & \textbf{\textcolor{blue}{0.181}}
& \textbf{\textcolor{red}{21.11}} & \textbf{\textcolor{blue}{0.806}} & \textbf{\textcolor{red}{0.181}} \\
\bottomrule
\bottomrule
\end{tabular}
\end{adjustbox}
\vspace{-5mm}
\end{table*}

Stage I: geometry pre-training.  
We train the geometry branch on dark images with photometric, gradient, depth, and normal constraints:
\begin{equation}
\begin{aligned}
\mathcal{L}_{\text{geo}}
&= \lambda_{\text{mse}}\mathcal{L}_{\text{mse}}^{\text{dark}}
+ \lambda_{\text{grad}}\mathcal{L}_{\text{grad}} \\
&\quad 
+ \lambda_{\text{depth}}\mathcal{L}_{\text{depth}}
+ \lambda_{\text{normal}}\mathcal{L}_{\text{normal}}.
\end{aligned}
\end{equation}
$\mathcal{L}_{\text{grad}}$ enforces edge consistency between grayscale gradients of predicted and target images,  
$\mathcal{L}_{\text{depth}}$ ensures depth smoothness guided by image edges,  
and $\mathcal{L}_{\text{normal}} = 1 - \langle \hat{\mathbf{n}}_{\text{pred}}, \mathbf{n}_{\text{gt}} \rangle$,  
where $\mathbf{n}_{\text{gt}}$ is estimated by StableNormal~\cite{ye2024stablenormal}.

Stage II: global enhancement.
In addition to photometric supervision on both dark and enhanced images,
we jointly optimize three style embeddings
$\{i\in\{\text{high},\,\text{low},\,\text{diff}\}\}$ with brightness correlation and domain regularization:
\begin{equation}
\begin{aligned}
\mathcal{L}_{\text{global}}
&= \lambda_{\text{style}}
   \sum_i \|s_\text{bright}^i - d_i\|_2^2   \\ 
&\quad + \lambda_{\text{lum}}\mathcal{L}_{\text{corr}}(\Delta\text{lum}, s_{\text{bright}}) + \lambda_{\text{dom}}\mathcal{L}_{\text{dom}}.
\end{aligned}
\end{equation}

Stage III: local refinement.
ARM estimates illumination, material, and view factors with predicted illumination decomposition from~\cite{zhang2019kindling} as supervision :
\begin{equation}
\begin{aligned}
\mathcal{L}_{\text{phys}} &=
\lambda_L\|A_L-\hat{I}_{\text{illum}}\|_1
+ \lambda_M\|A_M-\hat{I}_{\text{refl}}\|_1 \\
&\quad
+ \lambda_x\|A_X-\hat{I}_{\text{illum-diff}}\|_1.
\end{aligned}
\end{equation}

Across appearance stages, photometric, perceptual, and color-space consistency losses are applied for stable appearance optimization::
\begin{equation}
\begin{aligned}
\mathcal{L}_{\text{recon}} &=
\lambda_{\text{mse}}\big(
\mathcal{L}_{\text{mse}}^{\text{dark}}
+ \mathcal{L}_{\text{mse}}^{\text{enh}}
\big), \\[3pt]
\mathcal{L}_{\text{lpips}} &=
\lambda_{\text{lpips}}\big(
\mathcal{L}_{\text{lpips}}^{\text{dark}}
+ \alpha_{\text{enh}}\,
\mathcal{L}_{\text{lpips}}^{\text{enh}}
\big), \\[3pt]
\mathcal{L}_{\text{hsv}} &=
\lambda_{\text{hsv}}\big(
\mathcal{L}_h^{\text{circ-L1}}
+ \mathcal{L}_s^{\text{masked-L1}}
+ \mathcal{L}_v^{\text{L1}}
\big).
\end{aligned}
\end{equation}
The overall objective combines all terms:
\[
\mathcal{L} =
\mathcal{L}_{\text{geo}}
+ \mathcal{L}_{\text{global}}
+ \mathcal{L}_{\text{phys}}
+ \mathcal{L}_{\text{recon}}
+ \mathcal{L}_{\text{lpips}}
+ \mathcal{L}_{\text{hsv}}.
\]

\section{Experiment}
\subsection{Experiment Setup}

\begin{figure}[t]
  \centering
   \includegraphics[width=1.0\linewidth]{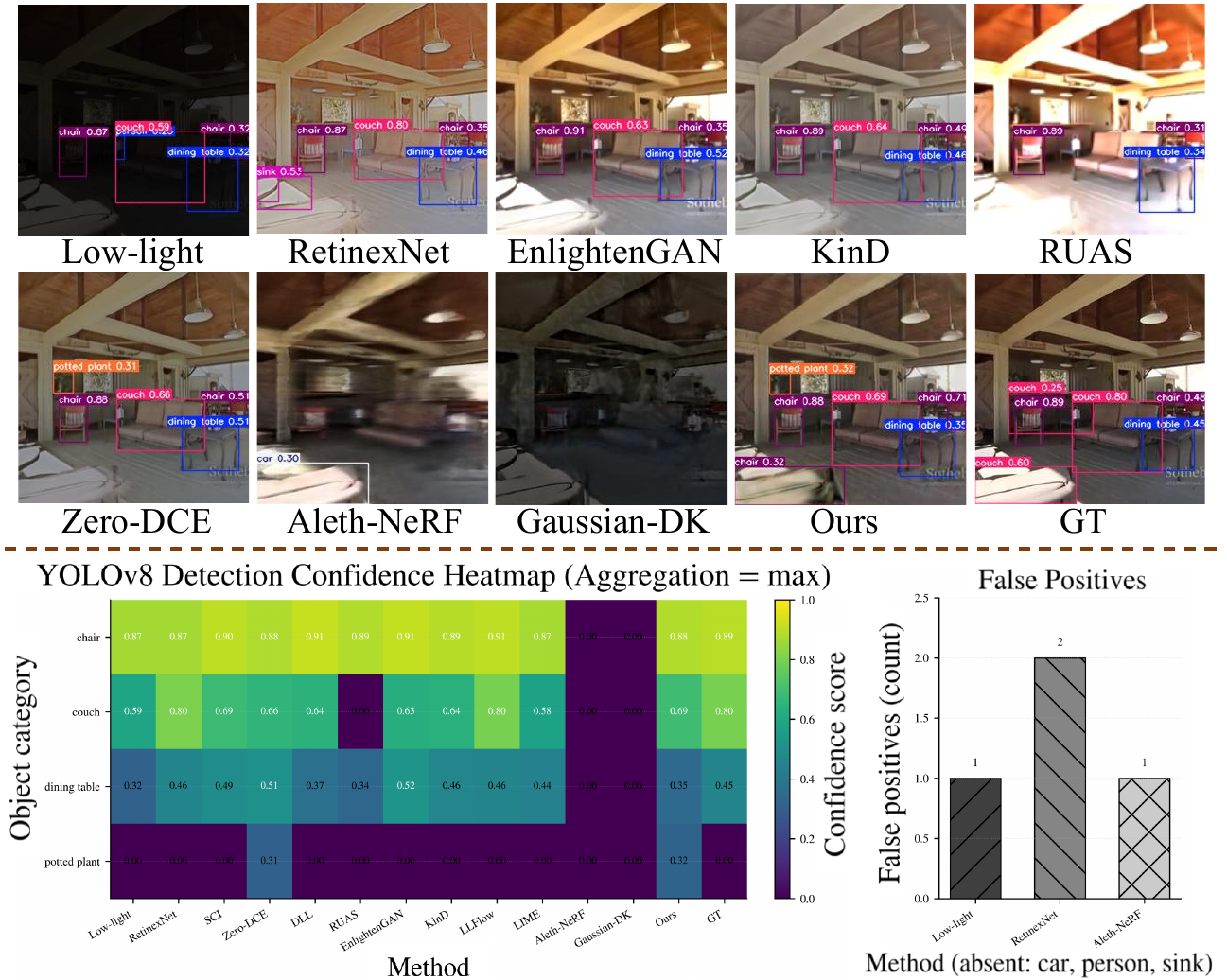}
   \vspace{-7mm}
   \caption{Object detection results on the RE10K dataset.}
   \label{fig:object}
   \vspace{-4mm}
\end{figure}



\noindent\textbf{Datasets.}
We primarily train and evaluate SplatBright on the large-scale indoor dataset \textbf{RE10K}~\cite{10.1145/3197517.3201323}, following the protocols of pixelSplat~\cite{charatan2024pixelsplat} and MVSplat~\cite{chen2024mvsplat}. RE10K contains $67{,}477$ training and $7{,}289$ testing scenes with per-frame camera intrinsics and extrinsics for multi-view supervision. To assess generalization, we evaluate on the cross-domain dataset \textbf{ACID}~\cite{liu2021infinite}, which provides $11{,}075$ training and $1{,}972$ testing aerial outdoor scenes with similar data configuration.
We also perform zero-shot evaluation on five standard low-light datasets (\textbf{NPE}~\cite{wang2013naturalness}, \textbf{MEF}~\cite{ma2015perceptual}, \textbf{LIME}~\cite{guo2016lime}, \textbf{DICM}~\cite{lee2013contrast}, \textbf{VV}\footnote{\url{https://sites.google.com/site/vonikakis/datasets}}). Since the lack camera poses, we set all poses to identity matrices during inference.
Additionally, we construct a real-world dataset \textbf{L3DS} with six scenes (five indoor, one outdoor). Each scene contains $25$-$35$ images captured by a tripod-mounted camera, with $2-3$ exposure levels per viewpoint. Images are downsampled from $3024\times3024$ to $400\times400$ or $800\times800$. Poses are estimated using COLMAP~\cite{schonberger2016structure} on normal images. For evaluation across all datasets, we use two context views and three novel target views located between them per scene.

\noindent\textbf{Implementation Details.}
SplatBright is implemented in PyTorch with a CUDA-based differentiable rasterizer.
All models are trained on $256\times256$ images for 120,000 iterations using Adam optimizer ($1.5\times10^{-4}$ initial learning rate) with cosine decay on a single A100 GPU.
We adopt a window size of $16^3$ for windowed 3D cross-attention (WCA).
Our three-stage training is scheduled to end at $5,000$, $25,000$, and $120,000$ iterations for geometry, global, and local optimization, respectively.
Training on RE10K takes approximately $20$ hours, and inference runs at $40$ FPS. 
More details are provided in the supplementary material.

\noindent\textbf{Metrics.}
We evaluate rendering quality using PSNR~\cite{huynh2008scope}, SSIM~\cite{wang2004image}, LPIPS~\cite{zhang2018unreasonable}, and the no-reference NIQE~\cite{mittal2012making} and MUSIQ~\cite{ke2021musiq} for perceptual realism.

\noindent\textbf{Baseline Methods.}
Since no prior work addresses our task under the same setting, we compare with the following approaches: (1) \textbf{original MVSplat}; (2) \textbf{SOTA 2D enhancement} (RetinexNet~\cite{Chen2018Retinex}, EnlightenGAN~\cite{jiang2021enlightengan}, KinD~\cite{zhang2019kindling}, LLFlow~\cite{wang2022low}, LIME~\cite{guo2016lime}, RUAS~\cite{liu2021retinex}, Zero-DCE~\cite{Zero-DCE}, DiffLL~\cite{wang2022low}, and SCI~\cite{ma2022toward}) with MVSplat pipelines: both 2D pre-process (\emph{method+MVSplat}) and 2D post-process (\emph{MVSplat+method});
(3) \textbf{3D Low-light reconstruction baselines}: Aleth-NeRF~\cite{cui2024aleth}, Gaussian-DK~\cite{ye2024gaussian}, which model illumination explicitly but require per-scene optimization, whereas our method enables direct feed-forward inference.

\subsection{Quantitative and qualitative results}

\begin{figure}[t]
  \centering
   \includegraphics[width=1.0\linewidth]{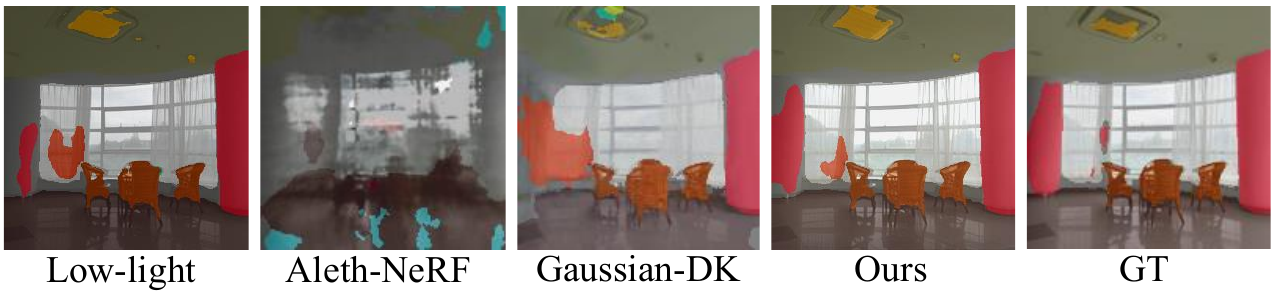}
   \vspace{-7mm}
   \caption{Visual comparison of segmentation performance among 3D-based methods on the L3DS dataset.}
   \label{fig:segformer}
   \vspace{-5.5mm}
\end{figure}

We construct paired data on the RE10K and ACID datasets following~\cref{sec:darkdata}. We train and evaluate our model on ~\textbf{RE10K}, with results shown in~\cref{fig:re10k} and~\cref{tab:re10k}. SplatBright achieves the best results, demonstrating clear advantages in low-light multi-view reconstruction. Best and second-best results are in red and blue. To assess cross-dataset generalization, we directly apply the RE10K-trained model to ~\textbf{ACID}. As shown in~\cref{fig:acid} and~\cref{tab:acid}, our method reliably estimates illumination and recovers realistic colors in unseen scenes, consistently achieving top performance.
We further evaluate on ~\textbf{VV, LIME, MEF, DICM, and NPE} using NIQE and MUSIQ. As shown in~\cref{tab:onlypic}, our method performs strongly on . Although some metrics on NPE and DICM are not strictly the best,~\cref{fig:onlypic} shows that our results provide the most natural brightness and smoothest textures. In contrast, 2D methods often produce blotchy textures or amplified noise, Aleth-NeRF shows limited brightening with blurry edges, and Gaussian-DK frequently fails under sparse views.
Finally, we evaluate on our self-captured ~\textbf{L3DS} dataset. As shown in~\cref{tab:L3DS} and~\cref{fig:L3DS}, SplatBright substantially outperforms all 2D and 3D baselines: 2D methods tend to over-sharpen, Aleth-NeRF yields blurry novel views, and Gaussian-DK suffers from color shifts.
Note that both Gaussian-DK and Aleth-NeRF are retrained per scene.

\subsection{Application to Downstream Vision Tasks}

\begin{figure}[t]
  \centering
   \includegraphics[width=1.0\linewidth]{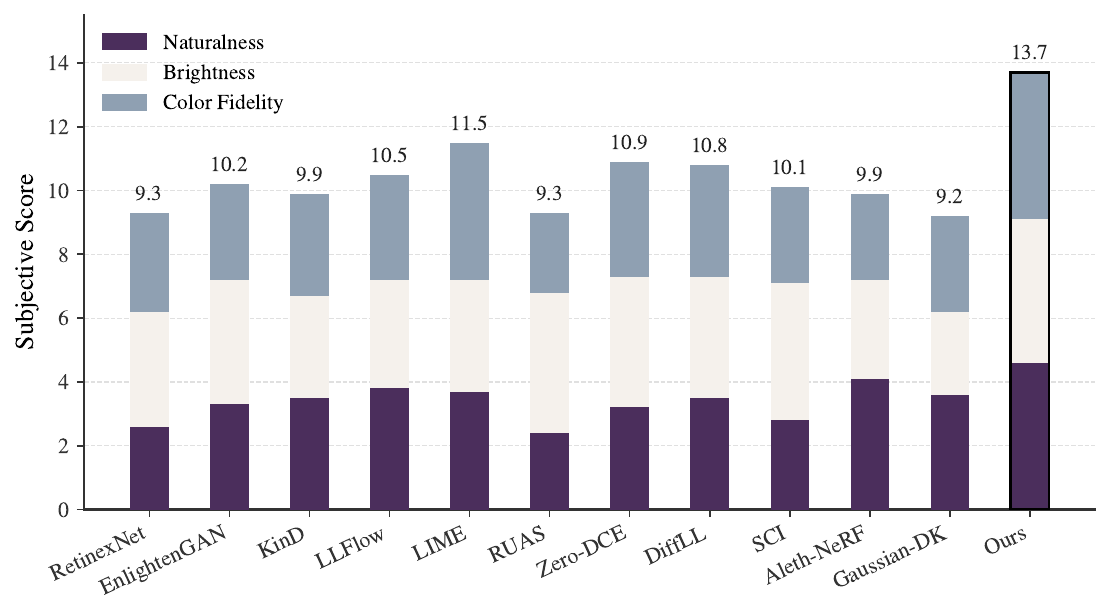}
   \vspace{-8mm}
   \caption{Comparison of subjective evaluation results on the NPE, DICM, MEF, LIME, and VV datasets.}
   \label{fig:subjective}
   \vspace{-6mm}
\end{figure}

We further evaluate the effectiveness of our method on downstream vision tasks. For object detection, we adopt YOLOv8~\cite{Jocher_Ultralytics_YOLO_2023} on the RE10K dataset. As shown in \cref{fig:object}, enhancement improves detection confidence in dark regions and reduces false positives. Notably, only our method and Zero-DCE correctly detect the “potted plant” missing from the ground truth, while Zero-DCE fails to detect all objects.
For semantic segmentation, we evaluate SegFormer~\cite{xie2021segformer} on our L3DS dataset. Our enhanced results produce more accurate segmentation than other 3D baselines in \cref{fig:segformer}, demonstrating that our approach not only improves visual quality but also benefits downstream perception tasks.

\subsection{Subjective Evaluation}

Since traditional quality metrics may not fully align with human perception~\cite{jiang2022single} and some datasets lack ground truth, we conduct a subjective evaluation to assess visual quality. Twenty volunteers rated enhancement results of 15 real low-light scenes from NPE, DICM, MEF, LIME, and VV in terms of naturalness, brightness, and color fidelity (1–5 scale). Our method achieves the highest overall scores in \cref{fig:subjective}, demonstrating superior perceptual quality.

\subsection{Ablation Studies}

\begin{figure}[t]
  \centering
   \includegraphics[width=1.0\linewidth]{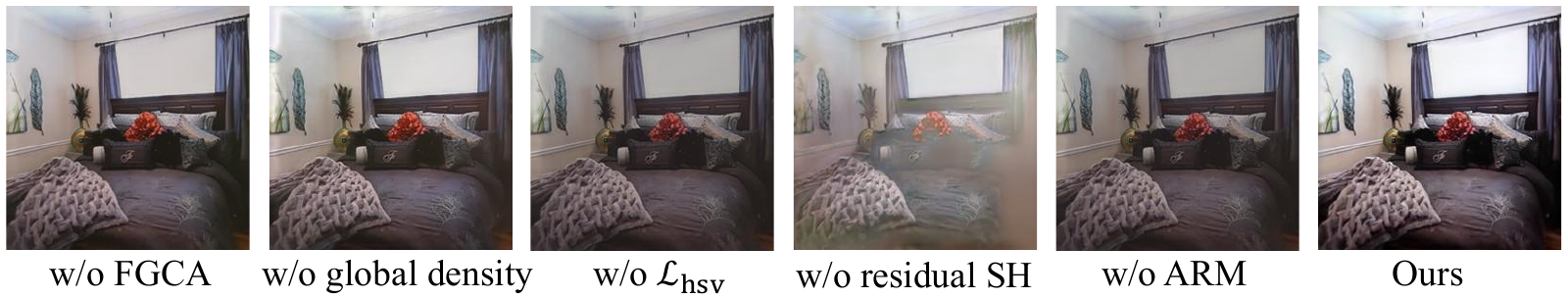}
   \vspace{-7mm}
   \caption{Visual comparison of ablation study on RE10K dataset.}
   \label{fig:ablation1}
   \vspace{-3mm}
\end{figure}

\begin{table}[t]
\caption{
Ablation study on different module and loss components.
}
\vspace{-2mm}
\label{tab:ablation}
\centering
\footnotesize
\renewcommand\arraystretch{0.9}
\begin{adjustbox}{max width=1\linewidth}
\begin{tabular}{l|c}
\toprule
\toprule
\multicolumn{2}{c}{\textbf{Module Ablation (PSNR$\uparrow$ / SSIM$\uparrow$ / LPIPS$\downarrow$)}} \\
\midrule
w/o physical dark data generation (use simplified model) & 19.95 / 0.781 / 0.216 \\
w/o global residual SH (use multiplicative SH)           & 19.01 / 0.767 / 0.231 \\
w/o global density optimization                          & 18.20 / 0.735 / 0.281 \\
w/o geo-freeze in appearance stage                       & 21.32 / 0.807 / 0.170 \\
w/o ARM                                                  & 20.93 / \textbf{\textcolor{blue}{0.813}} / 0.170 \\
w/o FGCA                                                 & \textbf{\textcolor{blue}{21.40}} / 0.808 / \textbf{\textcolor{blue}{0.170}} \\
\midrule
\textbf{Full model (ours)} & 
\textbf{\textcolor{red}{21.43}} /
\textbf{\textcolor{red}{0.815}} /
\textbf{\textcolor{red}{0.168}} \\
\midrule
\multicolumn{2}{c}{\textbf{Loss Ablation (PSNR$\uparrow$ / SSIM$\uparrow$ / LPIPS$\downarrow$)}} \\
\midrule
w/o $\mathcal{L}_{\text{hsv}}$        & 19.96 / 0.781 / 0.191 \\
w/o $\mathcal{L}_{\text{style}}$      & 21.28 / \textbf{\textcolor{blue}{0.809}} / 0.174 \\
w/o $\mathcal{L}_{\text{normal}}$     & \textbf{\textcolor{blue}{21.43}} / 0.804 / \textbf{\textcolor{blue}{0.169}} \\
\midrule
\textbf{Full model (ours)} & 
\textbf{\textcolor{red}{21.43}} /
\textbf{\textcolor{red}{0.815}} /
\textbf{\textcolor{red}{0.168}} \\
\bottomrule
\bottomrule
\end{tabular}
\end{adjustbox}
\vspace{-6mm}
\end{table}

Table~\ref{tab:ablation} summarizes the influence of each component. Replacing our physically driven dark-data modeling with a simplified gray-decay model yields a clear performance drop, demonstrating the importance of realistic illumination modeling. In ICM, switching from residual SH to a multiplicative formulation causes unstable brightness correction and overexposed regions in \cref{fig:ablation1}, while disabling global density optimization produces floating artifacts. In local appearance modeling, removing ARM degrades color fidelity and yields grayish outputs, whereas discarding FGCA disrupts multi-frequency aggregation and produces blurred textures (e.g., wall corners) with higher LPIPS. Regarding losses, removing $\mathcal{L}{\text{hsv}}$ or $\mathcal{L}{\text{style}}$ reduces color and contrast consistency, and \cref{fig:ablation2} shows that omitting $\mathcal{L}_{\text{normal}}$ weakens geometry–appearance decoupling, harming reconstruction. More details are provided in the supplementary material.

\label{sec:experiment}
\section{Conclusion}
We introduced \textbf{SplatBright}, a generalizable low-light 3D Gaussian reconstruction framework. With decoupled geometry–appearance modeling and physically guided illumination refinement, it achieves robust low-light reconstruction and consistently outperforms 2D and 3D baselines on public datasets and our self-captured multi-view data.

\clearpage
{
    \small
    \bibliographystyle{ieeenat_fullname}
    \bibliography{main}
}


\end{document}